\documentclass{article}
\usepackage{spconf,amsmath,graphicx,hyperref}
\usepackage{tikz}

\usepackage{graphicx}

\usepackage{tcolorbox}
\usepackage{multirow}
\usepackage{booktabs} 
\usepackage{colortbl}

\usepackage{enumitem}
\title{Beyond Global Emotion: Fine-Grained Emotional Speech Synthesis with Dynamic Word-Level Modulation}
\name{Sirui Wang, Andong Chen, Tiejun Zhao}

\address{Harbin Institute of Technology, Harbin, China}
\begin{document}
\ninept

%
\maketitle

\begingroup
\begin{tikzpicture}[overlay, remember picture]
    \node[anchor=north, yshift=-1.5cm] at (current page.north) { 
        \begin{minipage}{\textwidth} 
            \centering 
            \textcolor{red}{\small © 2026 IEEE. Personal use of this material is permitted. Permission from IEEE must be obtained for all other uses, in any current or future media, including reprinting/republishing this material for advertising or promotional purposes, creating new collective works, for resale or redistribution to servers or lists, or reuse of any copyrighted component of this work in other works.}
        \end{minipage}
    };
\end{tikzpicture}
\endgroup

\begin{abstract}
Emotional text-to-speech (E-TTS) is central to creating natural and trustworthy human–computer interaction. Existing systems typically rely on sentence-level control through predefined labels, reference audio, or natural language prompts. While effective for global emotion expression, these approaches fail to capture dynamic shifts within a sentence. To address this limitation, we introduce Emo-FiLM, a fine-grained emotion modeling framework for LLM-based TTS. Emo-FiLM aligns frame-level features from emotion2vec to words to obtain word-level emotion annotations, and maps them through a Feature-wise Linear Modulation (FiLM) layer, enabling word-level emotion control by directly modulating text embeddings. To support evaluation, we construct the Fine-grained Emotion Dynamics Dataset (FEDD) with detailed annotations of emotional transitions. Experiments show that Emo-FiLM outperforms existing approaches on both global and fine-grained tasks, demonstrating its effectiveness and generality for expressive speech synthesis.

\end{abstract}
\section{Introduction}

Emotional text-to-speech (E-TTS) plays a key role in building natural and trustworthy human-computer interaction~\cite{DBLP:conf/icassp/LiCHCFPH25,tan2021survey,cui2021emovie}. It has been widely used in voice assistants, virtual characters, and multimodal systems to enhance user engagement and immersion~\cite{chang2023importance,DBLP:conf/interspeech/GudmalwarSAWS24,chan2024human}.

Most existing E-TTS methods control emotion at the sentence level. Current methods can be grouped into three main categories. The first is predefined emotion labels~\cite{10800745,10423864,DBLP:conf/interspeech/ChoOKLL24}, where models are trained with discrete labels such as happy, sad, or angry. These labels are used to supervise the learning of emotional embeddings or classifiers. The second is reference audio-based control~\cite{DBLP:conf/nips/Huang0LCZ22,jawaid2024style,patra2013saras}, which extracts style or emotion embeddings from reference audio for zero-shot emotion transfer. This allows cross-speaker and cross-domain transfer but depends on whole-sentence reference audio. The third is natural language prompts with Large Language Models TTS (LLM-TTS)~\cite{DBLP:journals/corr/abs-2412-10117,bott2024controlling,du2024cosyvoice}, where natural language descriptions are prepended to the text along with special tokens (e.g., [laughter], [breath]) to specify emotion, speed, role, or accent. While these methods offer flexibility, they rely on global control signals and cannot capture changes in emotion over time. For example, a sentence may begin with a surprised tone and shift to joy, but such dynamic variations are hard to express using global controls.

To address this limitation, we propose Emo-FiLM, a fine-grained emotion modeling method based on Feature-wise Linear Modulation (FiLM). Our approach first uses an emotion2vec model to extract frame-level emotion features and aligns them to generate word-level dynamic emotion annotations. We then introduce an emotion modulation layer (E-FiLM) into an LLM-TTS framework. This layer transforms word-level emotion signals into dimension-wise scale and shift parameters that modulate the text embeddings. 

\begin{figure}
    \centering
    \includegraphics[width=\linewidth]{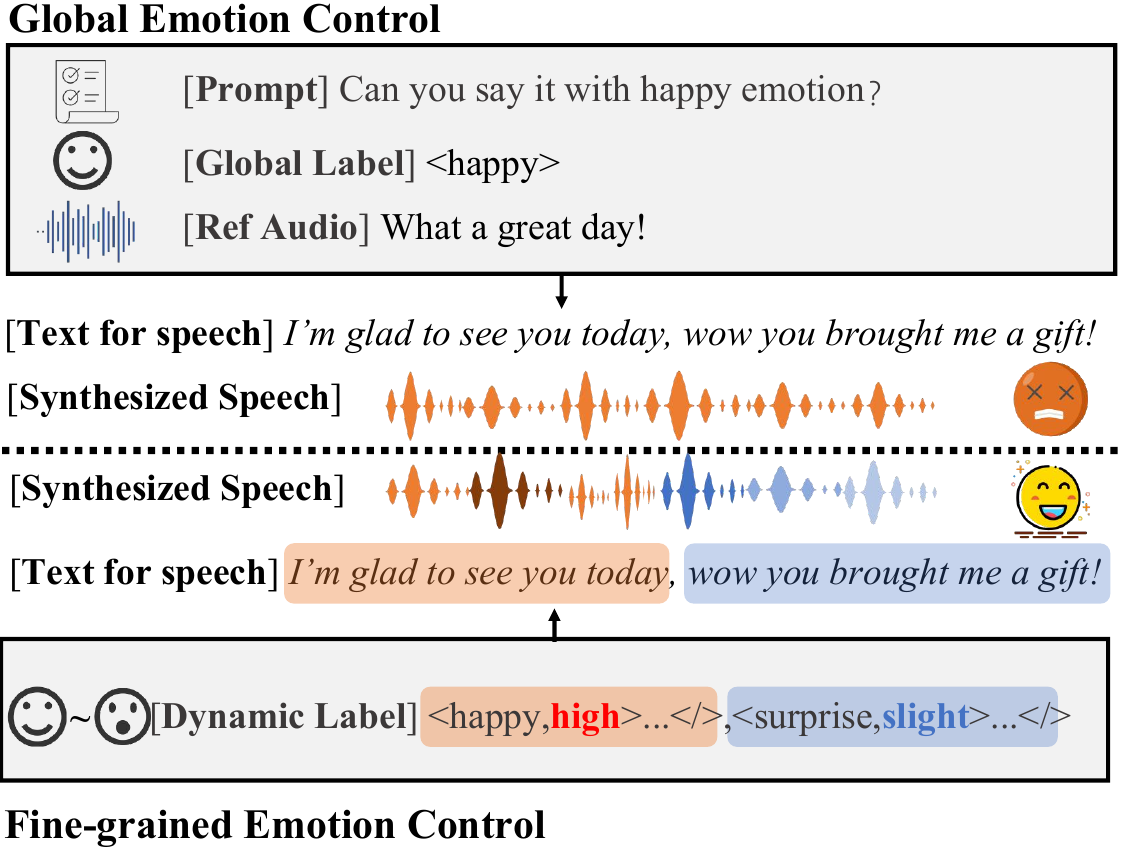}
\caption{Global emotion control relies on sentence-level signals, while our method enables fine-grained word-level modulation for dynamic emotion shifts.}
    \label{fig:intro}
\end{figure}
As shown in Figure~\ref{fig:intro}, unlike existing approaches that rely on global emotion control, our method achieves fine-grained word-level modulation, enabling the modeling of dynamic emotional transitions that global methods cannot capture. To evaluate the model’s ability in fine-grained emotion control, we construct a new dataset named Fine-grained Emotion Dynamics Dataset (FEDD), which fills the gap in existing benchmarks lacking detailed emotion annotations.

\begin{figure*}[!t]
    \centering
    \includegraphics[width=0.9\linewidth]{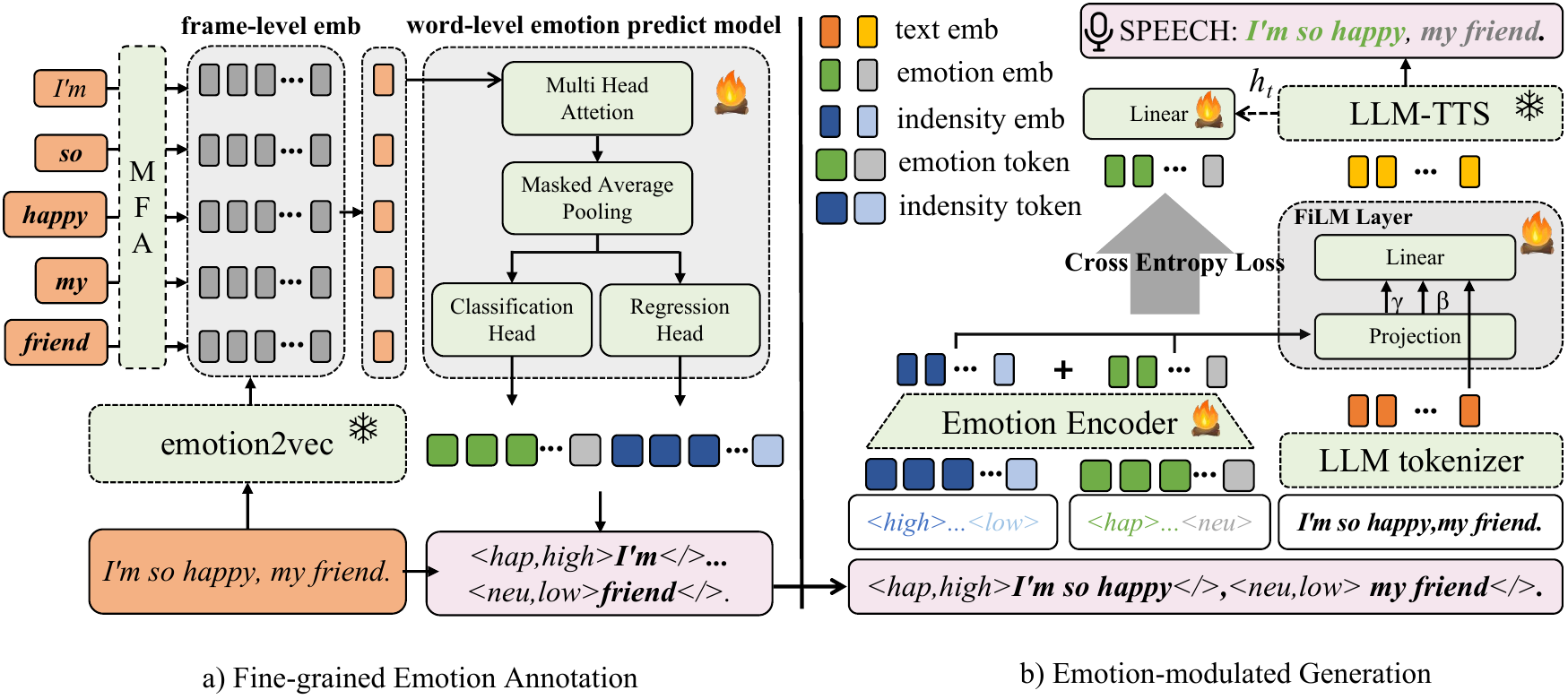}
\caption{Overview of the proposed \textbf{Emo-FiLM} framework. 
(a) Fine-grained emotion annotation aligns frame-level features with text to predict emotion category and intensity. 
(b) Emotion-modulated generation uses an Emotion Encoder and FiLM layer to inject these signals into LLM-TTS for fine-grained controllable synthesis.}
    \label{fig:method}
\end{figure*}

Compared to global emotion control methods, our proposed approach achieves better results on both the ESD global emotion synthesis task and the FEDD dynamic emotion task. On the global task, Emo-FiLM achieves higher emotion similarity (Emo SIM). On the dynamic task, it significantly outperforms baselines in emotion dynamic matching (DTW) and subjective similarity (EMOS). Further analysis shows that combining frame-level emotion features with FiLM-based modulation improves the model’s ability to model emotion shifts within a sentence. Overall, our results demonstrate the effectiveness and generality of the proposed method for fine-grained controllable emotional TTS. To the best of our knowledge, this is the first work to leverage LLM-TTS for dynamic control of fine-grained emotions in speech synthesis.


\section{Methodology}
\label{sec:method}

We propose Emo-FiLM for word-level controllable emotional speech synthesis. As shown in Figure \ref{fig:method}, it consists of two modules: (1) Fine-grained Emotion Annotation, where emotion2vec extracts frame-level features and aligns them with words to form word-level trajectories; (2) Emotion-modulated Generation, where an E-FiLM layer maps these trajectories into scale and shift parameters to modulate text states, enabling dynamic prosody and emotion control.

\subsection{Fine-grained Emotion Annotation}

Most existing emotional speech datasets are annotated at the sentence level, which limits their ability to support fine-grained modeling of dynamic emotions within a sentence. To obtain more fine-grained text--speech emotion alignment, we extract frame-level emotion features using emotion2vec and further infer word-level emotion representations. emotion2vec is a self-supervised speech emotion recognition model capable of capturing high-dimensional emotion-related features from speech.

As shown in Figure \ref{fig:method} (a), 
we train a lightweight Transformer model with multi-head self-attention layers, feed-forward networks, and residual connections to model contextual dependencies of frame sequences and generate enhanced representations. Specifically, we first use the \textbf{Montreal Forced Aligner (MFA)} to align speech with transcribed text, obtaining frame-level emotion feature sequences corresponding to each word. Then, masked average pooling is applied to aggregate each sequence into a fixed-dimensional vector, which serves as the word-level emotion representation. This representation is passed through two parallel output heads: a classification head for predicting discrete emotion categories, and a regression head for estimating continuous emotion intensity (normalized to the range $[0,1]$). For a frame-level emotion feature sequence $\mathbf{x}_{1:T}$ corresponding to a word, the prediction targets include the discrete category $\hat{y}^{\text{class}}$ and the continuous intensity $\hat{y}^{\text{dim}}$:

\begin{equation}
\hat{y}^{\text{class}}, \hat{y}^{\text{dim}} = f_\theta(\mathbf{x}_{1:T}),
\end{equation}
where $\mathbf{x}_{1:T}$ denotes the frame-level emotion feature sequence aligned with a word; $\hat{y}^{\text{class}}$ is the predicted emotion category; and $\hat{y}^{\text{dim}}$ is the predicted emotion intensity. During training, the model jointly optimizes classification and regression tasks for fine-grained emotion modeling. The loss function is defined as:

\begin{equation}
\mathcal{L} = \lambda_{\text{cls}} \cdot \mathcal{L}_{\text{CE}}(\hat{y}^{\text{class}}, y^{\text{class}})
+ \lambda_{\text{reg}} \cdot \mathcal{L}_{\text{MSE}}(\hat{y}^{\text{dim}}, y^{\text{dim}}),
\end{equation}
where $\mathcal{L}_{\text{CE}}$ and $\mathcal{L}_{\text{MSE}}$ are the classification and regression losses weighted by $\lambda_{\text{cls}}$ and $\lambda_{\text{reg}}$, and the model infers word-level dynamic annotations from frame-level features to provide fine-grained control for speech synthesis.

\begin{table*}[!t]\small
\centering
\resizebox{\linewidth}{!}{%
\begin{tabular}{llcccccc}
\toprule
\textbf{Model} & \textbf{Emotion} & \textbf{Dataset} & \textbf{Emo SIM(\%)} & \textbf{DTW} & \textbf{WER(\%)}& \textbf{EMOS}& \textbf{NMOS}\\
\midrule
EmoSpeech     & Label               & ESD   & 98.25 &    47.34& 7.92 &   4.09&   3.93\\
GenerSpeech   & Audio               & ESD   &   97.84&    42.68& 12.35 &   3.72&   3.81\\
CosyVoice2    & Prompt              & ESD   & 98.73 & 27.48& 6.21&   4.07&   4.19\\
Emo-FiLM& Global Label        & ESD   & 98.78 & 23.98&  3.12&   4.13&   4.23\\
\midrule
EmoSpeech     & Label               & FEDD&   98.33&    59.89&  8.04&   3.99&   3.96\\
GenerSpeech   & Audio               & FEDD&   98.17&    65.63&  9.58&   3.62&   3.82\\
CosyVoice2    & Prompt              & FEDD& 99.13& 54.57&  9.93&   3.84&   4.17\\
Emo-FiLM& Fine-grained Label  & FEDD& 99.32 & 49.62&  7.32&   4.19&   4.23\\
\bottomrule
\end{tabular}}
\caption{Performance comparison of different emotional TTS models on ESD and FEDD datasets.}
\label{tab:emo_tts_results}
\end{table*}

\subsection{Emotion-modulated Generation}
To achieve fine-grained emotional control, we introduce the \textbf{Emotion Feature-wise Linear Modulation (E-FiLM)} module into the pretrained LLM-TTS framework. As shown in Figure \ref{fig:method} (b), this module consists of three components: an \textit{Emotion Encoder}, a \textit{FiLM Layer}, and an \textit{Emotion Supervision Mechanism}. The emotion encoder maps discrete emotion categories and continuous intensity labels into dense vector representations, which are fused with text embeddings to obtain unified emotional feature sequences. The FiLM layer then projects these emotion features into dimension-wise scaling factors $\gamma$ and shifting factors $\beta$, and applies affine modulation to the text hidden representations $\mathbf{h}_{\text{text}}$:

\begin{equation}
\tilde{\mathbf{h}}_{\text{text}} = \gamma \odot \mathbf{h}_{\text{text}} + \beta ,
\end{equation}
where $\odot$ denotes element-wise multiplication. This operation allows the text representation to encode appropriate emotional signals at the feature dimension level, thereby enabling controllable prosody and dynamic emotional variation during speech synthesis.

During training, the model adopts a multi-task learning strategy to jointly optimize speech generation and emotion classification objectives. The speech generation task employs a label-smoothed cross-entropy loss to predict the next speech token, preserving the synthesis ability of the pretrained model:

\begin{equation}
\mathcal{L}_{\text{TTS}} = -\frac{1}{M} \sum_{i=1}^{B} \sum_{t=1}^{L_i} \sum_{k=1}^{V} q(y_{i,t},k)\log p_{i,t}(k),
\end{equation}

while the emotion classification task projects the hidden state at each time step into the emotion category space and applies a step-wise cross-entropy loss:

\begin{equation}
\mathcal{L}_{\text{emo}} = -\frac{1}{N} \sum_{i=1}^{B} \sum_{t=1}^{T_i} \log p_{i,t}(y_{i,t}).
\end{equation}

The overall training objective is a weighted combination of the two losses:

\begin{equation}
\mathcal{L} = \mathcal{L}_{\text{TTS}} + \lambda \mathcal{L}_{\text{emo}},
\end{equation}

where $\lambda$ is a balancing coefficient.




During inference, the input text is first jointly encoded with word-level emotional features, which are modulated by the FiLM layer before being fed into the LLM-TTS decoder. The decoder then generates the speech token sequence in an autoregressive manner. This process ensures that the synthesized speech can exhibit natural emotional dynamics and prosodic variations according to the word-level control signals.

\section{Experiments}
\label{sec:exp}

\subsection{Datasets}
To evaluate the model’s performance on both global and fine-grained emotional synthesis, we conduct experiments on two test sets.
For the global task, we build a test set from the English portion of ESD~\cite{zhou2022emotional}, covering 10 speakers and 5 emotions (Angry, Happy, Sad, Surprise, Neutral), with 30 utterances per category, totaling 1,500 samples.
For the fine-grained task, we construct the FEDD dataset, which contains 1,000 utterances with emotional shifts across 5 speakers and 5 emotions. Among them, 500 are mild transitions generated by natural language instructions, while 500 are strong transitions created by concatenating segments with different emotions from the same speaker. Global emotion labels are automatically assigned using emotion2vec-plus-large. This dataset is designed to assess the model’s ability to capture and generate emotional transitions.

\subsection{Model Implementation and Training}
We use the global emotion labels from IEMOCAP\cite{busso2008iemocap} and ESD to pseudo-label words and train a word-level emotion prediction model, which maps frame-level features to word-level representations for generating fine-grained speech–text pairs. Based on this, Emo-FiLM is built on the frozen pre-trained CosyVoice2 framework, where only the E-FiLM module and the decoding head are optimized, and integrates Flow Matching with HiFi-GAN~\cite{kong2020hifi} to achieve word-level emotion-controllable speech synthesis. Training is performed with the Adam optimizer, a batch size of 4, and 5 epochs.

\subsection{Baselines}
To comprehensively evaluate the advantages of Emo-FiLM in emotion control and speech generation, we compare it with three representative baselines: the label-based fastspeech2-EmoSpeech~\cite{10800745}, the reference-audio-based GenerSpeech~\cite{DBLP:conf/nips/Huang0LCZ22}, and the natural-language-instruction-based CosyVoice2~\cite{DBLP:journals/corr/abs-2412-10117}. For EmoSpeech and CosyVoice2, we use test data with global emotion labels, while for GenerSpeech, reference audio with the same emotion and speaker as the ground truth is provided.

\subsection{Evaluation Metrics}
We evaluate Emo-FiLM and baseline models using both objective and subjective metrics. Objective metrics include Emo SIM~\cite{DBLP:conf/acl/MaZYLGZ024} to measure emotion similarity between synthesized and real speech, —note that Emo SIM averages frame-level emotion vectors, which may partially obscure dynamic emotional information. To mitigate this limitation, we further introduce Dynamic Time Warping (DTW) to provide an auxiliary assessment of emotion similarity.  WER is computed with Whisper-Large-v3~\cite{radford2023robust} to evaluate intelligibility. Subjective metrics include Mean Emotion Similarity Opinion Score (EMOS) and the Mean Opinion Score of Speech Naturalness (NMOS)~\cite{DBLP:conf/icassp/LiCHCFPH25}. EMOS measures the perceived similarity of listeners between the emotion conveyed by the synthesized speech and the target emotion, while NMOS quantifies the perceived naturalness of the speech.

\section{Results and Discussion}

\subsection{Main Results}
The main experiment aims to evaluate Emo-FiLM against representative emotional TTS baselines under both global and word-level control tasks.  As shown in Table~\ref{tab:emo_tts_results}, results on the ESD dataset show that Emo-FiLM achieves the best or competitive scores across all metrics, including a 12.7\% relative improvement in DTW over CosyVoice2 and a low WER of 3.12\%, indicating strong emotional expressiveness while maintaining intelligibility. On the more challenging FEDD dataset, Emo-FiLM further surpasses baselines with a 9.1\% DTW gain and higher subjective ratings in both emotion similarity and naturalness. These findings confirm that the proposed feature-wise modulation approach effectively captures dynamic emotions, and demonstrate that Emo-FiLM delivers robust performance in both global and fine-grained emotional speech synthesis.

\subsection{Ablation Study}
The goal of this experiment is to assess the effectiveness of different components in Emo-FiLM. As shown in Table~\ref{tab:emo_ablation}, training with only global-level labels significantly increases DTW on FEDD from 49.6 to 52.7, while removing word-level data causes the most severe degradation, with DTW rising to 134.0, clearly showing the necessity of fine-grained supervision. Removing the auxiliary emotion loss also leads to large drops (DTW 73.9 vs. 49.6), further confirming the benefit of explicit emotion classification. Finally, replacing the FiLM Layer with simple addition severely hurts performance (DTW 70.5 vs. 49.6), highlighting the importance of nonlinear affine modulation in capturing nuanced emotional variations. Overall, all ablations consistently degrade performance across both ESD and FEDD, thereby validating that each design choice is indeed critical for the overall effectiveness of Emo-FiLM.

\begin{table}[!t]
\centering
\resizebox{\linewidth}{!}{%
\begin{tabular}{lcccc}
\toprule
\textbf{Model} & \multicolumn{2}{c}{\textbf{ESD}} & \multicolumn{2}{c}{\textbf{FEDD}} \\
\cmidrule(lr){2-3} \cmidrule(lr){4-5}
 & Emo SIM & DTW & Emo SIM & DTW \\
\midrule
Emo-FiLM& 98.78 & 23.98& 99.32 & 49.62\\
\ \ - Global Level Data Tuning & 98.45 & 30.08& 99.20 & 52.72\\
\ \ - Word Level Data Tuning & 98.45 & 34.00& 95.28 & 133.97\\
\ \ - Word Level Data Tuning & 98.58 & 25.96& 98.99 & 55.91\\
\ \ - Emo Loss & 98.26 & 34.36& 98.83 & 73.96\\
\bottomrule
\end{tabular}}
\caption{Ablation study results on ESD and FEDD datasets.}
\label{tab:emo_ablation}
\end{table}

\subsection{Per-Emotion Classification Accuracy Analysis}
\begin{figure}[!ht]
    \centering
    \includegraphics[width=1.0\linewidth]{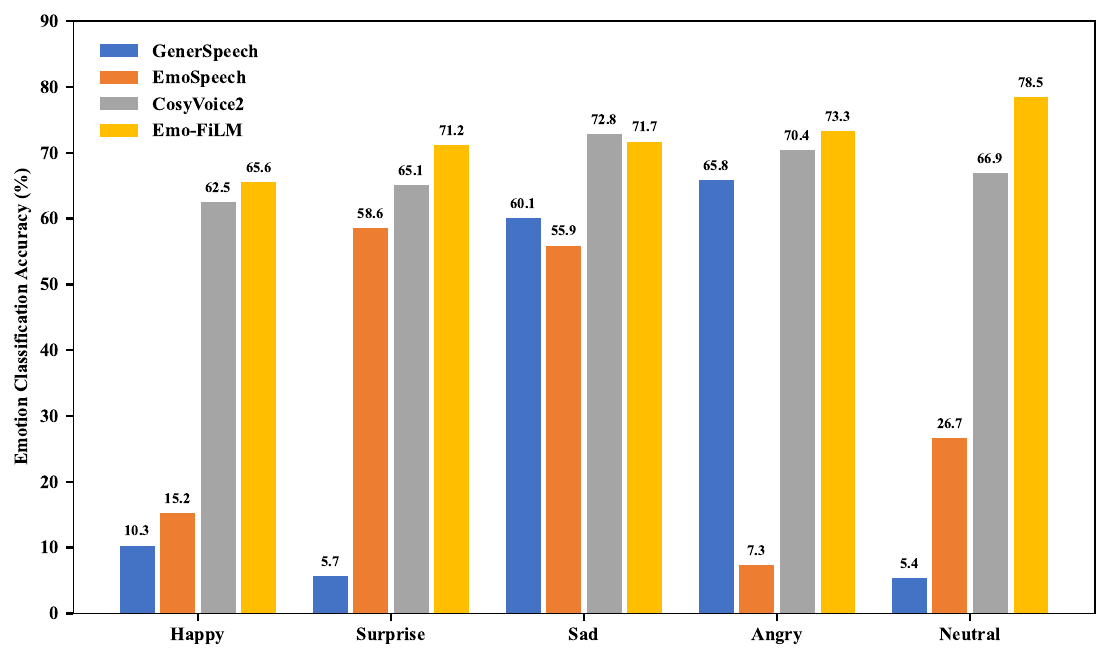}
\caption{Per-emotion classification accuracy on the ESD dataset.}
    \label{fig:recipe}
\end{figure}

The purpose of this experiment is to further analyze the emotion expressiveness of each model by comparing their classification accuracy across emotion categories on the ESD dataset. As shown in Figure~\ref{fig:recipe}, Emo-FiLM achieves the highest accuracy in Happy (65.6\%), Surprise (71.2\%), Angry (70.4\%), and Neutral (78.5\%), consistently outperforming the baselines. For example, on the Neutral class, Emo-FiLM improves over CosyVoice2 (69.8\%) and EmoSpeech (26.7\%) by a large margin, while in Happy it surpasses GenerSpeech (10.3\%) and EmoSpeech (18.3\%) by over 47\% absolute gain. These results demonstrate that Emo-FiLM delivers more precise and robust emotion rendering across multiple key categories, highlighting its superior discriminative power and expressiveness in emotional speech synthesis.

\subsection{Case Study: Visualization of Emotional Dynamics}
\begin{figure}[!ht]
    \centering
    \includegraphics[width=1.0\linewidth]{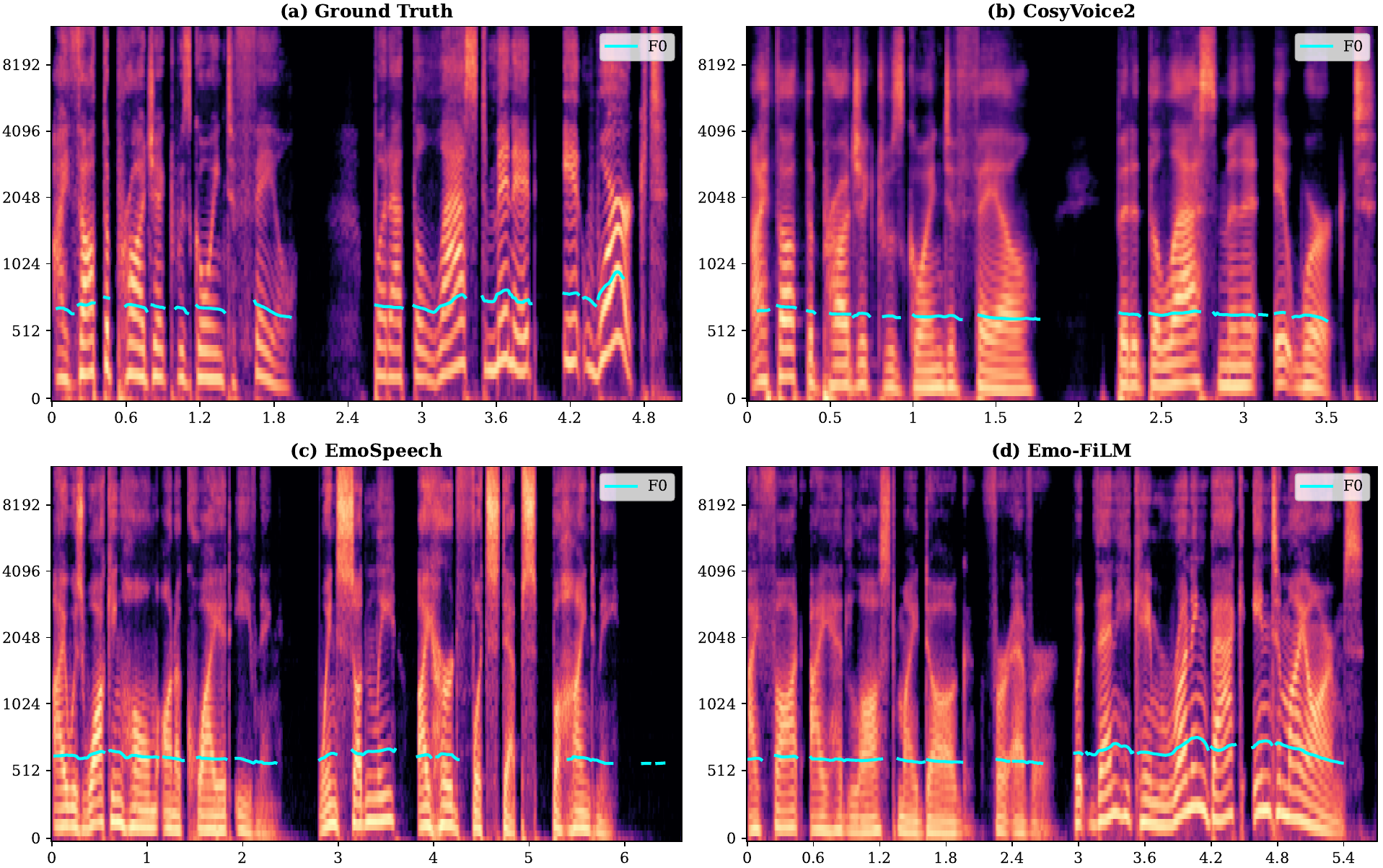}
    \caption{Visualization of mel-spectrograms and pitch (F0) contours for the utterance “I thought the project was going to fail, but at the last minute it worked!”, which contains emotion transitions.}
    \label{fig:case}
\end{figure}

To intuitively evaluate the fine-grained emotion control of Emo-FiLM, we visualize the mel-spectrograms and pitch (F0) contours of an utterance with emotion transitions. As shown in Figure~\ref{fig:case}, CosyVoice2 and EmoSpeech produce nearly flat F0 trajectories, failing to capture local prosodic variations such as pitch rises or falls around emotion boundaries. In contrast, Emo-FiLM generates F0 contours that closely follow the ground truth, reproducing both the global trend and local fluctuations of emotional dynamics. This indicates that our model can not only preserve the overall prosody but also capture subtle shifts in intonation, stress, and rhythm that correspond to fine-grained emotional changes. Such improvements highlight Emo-FiLM’s capability to deliver speech with richer expressiveness and more natural emotional transitions, providing direct acoustic evidence of its effectiveness in fine-grained emotion modeling.

\section{Conclusion}
In this work, we proposed Emo-FiLM, a word-level controllable framework for fine-grained emotional speech synthesis. By aligning frame-level features with words to obtain word-level emotion annotations and applying FiLM-based modulation, our method enables dynamic emotion control beyond global signals. Experiments on both global and fine-grained tasks show consistent improvements in emotion similarity, dynamic alignment, and classification accuracy over strong baselines. These results confirm the effectiveness and generality of Emo-FiLM, providing a promising direction for expressive and trustworthy human–computer interaction.

\vfill\pagebreak




\bibliographystyle{IEEEbib}
\bibliography{main}

\appendix




\end{document}